\documentclass[letterpaper]{article} % DO NOT CHANGE THIS
\usepackage{aaai2027}  % DO NOT CHANGE THIS
\usepackage[hyphens]{url}  % DO NOT CHANGE THIS
\usepackage{graphicx} % DO NOT CHANGE THIS
\usepackage{natbib}  % DO NOT CHANGE THIS AND DO NOT ADD ANY OPTIONS TO IT
\usepackage{caption} % DO NOT CHANGE THIS AND DO NOT ADD ANY OPTIONS TO IT
\nocopyright
\usepackage{algorithm}
\usepackage{algorithmic}

\usepackage{newfloat}
\usepackage{listings}
\DeclareCaptionStyle{ruled}{labelfont=normalfont,labelsep=colon,strut=off} % DO NOT CHANGE THIS
\floatstyle{ruled}
\newfloat{listing}{tb}{lst}{}
\floatname{listing}{Listing}

\usepackage{booktabs}

\usepackage{amsmath}
\usepackage{amssymb}
\usepackage{amsthm}

\newcommand{\method}{TASN}
\newcommand{\methodfull}{Timestep-Adaptive Spectral Nullification}
\newcommand{\Wov}{W_{\mathrm{OV}}}
\newcommand{\Sc}{\mathcal{S}_c}
\newcommand{\R}{\mathbb{R}}

\newtheorem{theorem}{Theorem}

\usepackage[table]{xcolor}
\definecolor{traincolor}{HTML}{E8E8E8}
\definecolor{closedcolor}{HTML}{D6EAF8}
\definecolor{infercolor}{HTML}{D5F5E3}
\definecolor{ourscolor}{HTML}{FCF3CF}
\definecolor{placeholdercolor}{HTML}{FFF3CD}

\usepackage{multirow}

\title{Not All Steps Matter: Temporally Adaptive Spectral Control\\ for Concept Removal in Diffusion Models}
\author{
    Chanaka Thushitha Bandara\textsuperscript{\rm 1,*},
    Lakmali Nadeesha Kumari\textsuperscript{\rm 2,*},
    Sen-Ching Samson Cheung\textsuperscript{\rm 2,3}
}
\affiliations{
    \textsuperscript{\rm 1}University of Delaware, Newark, DE 19716, USA\\
    \textsuperscript{\rm 2}University of California, Davis, CA 95616, USA\\
    \textsuperscript{\rm 3}University of Kentucky, Lexington, KY 40506, USA\\
    chanaka@udel.edu, lrathukohe@ucdavis.edu, sen-ching.cheung@uky.edu
}

\begin{document}

\maketitle

\begingroup
\renewcommand{\thefootnote}{*}
\footnotetext{Equal contribution.}
\endgroup

\begin{abstract}
As text-to-image diffusion models see widespread deployment, removing unsafe or undesirable concepts without retraining has become a pressing need. Spectral methods are attractive because they are training-free and reversible, yet current variants apply the same output-value (OV) circuit intervention at every denoising step, assuming concept information is temporally uniform. This wastes intervention where concept energy is low while degrading image quality, an inefficiency that compounds across concept types. Through timestep-resolved spectral analysis, concept-specific energy is found to follow a consistent temporal hierarchy: \emph{content} concepts decay fastest, \emph{object} concepts are intermediate, and \emph{style} concepts persist longest. This paper proposes \methodfull{} (\method{}), which adapts suppression strength and spectral components to the concept energy profile. Cross-attention at each timestep is decomposed into concept-aligned and concept-orthogonal components, and it is proved that when the orthogonal component is approximately stable, the collateral-minimizing schedule sets suppression proportional to concept energy. \method{} implements this with a timestep-adaptive spectral subset tracking only active concept directions, strictly generalizing uniform Spectral Nullification (SN). On five red-teaming benchmarks, \method{} lowers attack success rate below uniform SN while recovering the majority of its quality loss; ablations confirm gains arise from temporal targeting rather than reduced suppression, generalizing across concept categories and architectures.
\end{abstract}


% ══════════════════════════════════════════════════════════
% 1. INTRODUCTION
% ══════════════════════════════════════════════════════════

\section{Introduction}

Text-to-image diffusion models~\citep{ho2020denoising, rombach2022high,saharia2022photorealistic} generate images through an iterative denoising process, where cross-attention layers inject textual information into the visual latent representation at each step. Recent mechanistic analyses of diffusion-model have decomposed cross-attention into query-key (QK) and output-value (OV) circuits~\citep{elhage2021mathematical}, showing that semantic concepts can be localized in low-dimensional spectral subspaces of OV transformations~\citep{bae2026mechanistic}. This observation enables training-free concept removal: by identifying concept-specific singular directions and suppressing them during inference, one can reduce undesirable content without retraining or permanently modifying the model.

\begin{figure}[!t]
    \centering
    \includegraphics[width=\columnwidth]{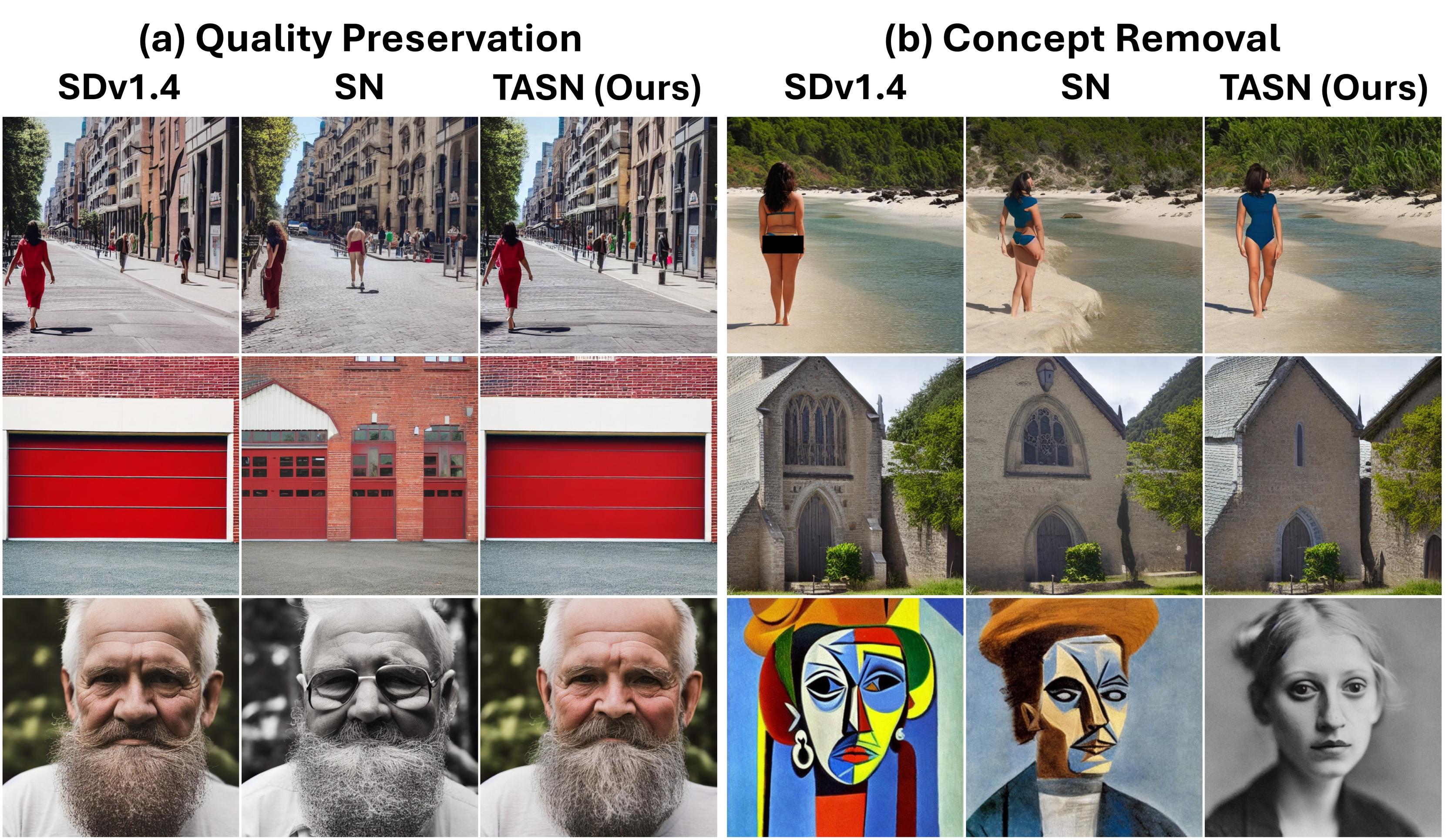}
    \caption{Quality preservation (a) and concept removal (b) across content, object, and style concepts (rows 1–3, respectively). Benign prompts: woman on a city street, red brick fire station, bearded man portrait. Concept-invoking prompts: woman in bikini on
beach, countryside church, Picasso
cubist face.}
\label{fig:qualitative}
\end{figure}

However, existing spectral methods make a strong temporal assumption. They compute a concept signature once and apply the same intervention at every denoising step. This treats concept information as if it were uniformly active throughout generation. In practice, denoising is a dynamic process: early steps often establish semantic layout, while later steps refine texture, color, and visual fidelity. Therefore, a spectral direction that is important for removing a concept at one timestep may carry mostly general visual information at another. Uniform nullification can consequently waste intervention at low-concept-energy steps and introduce unnecessary quality degradation
(see Figure~\ref{fig:qualitative}, where uniform SN visibly degrades
colors, layouts, and fine details on benign prompts).

The central question is: \emph{how does the spectral footprint of a concept evolve across the denoising trajectory, and can this temporal structure be used for more precise concept removal?} To answer this, a timestep-resolved spectral analysis of OV circuits is introduced. Instead of measuring only which spectral components encode a concept, this analysis measures when those components are active during generation. Across content, object, and style concepts, a consistent temporal hierarchy emerges: content concepts (e.g., nudity) decay fastest, object concepts (e.g., church) occupy an intermediate regime, and style concepts (e.g., Picasso) persist the longest.

The contributions are threefold. First, a timestep-resolved
spectral analysis is introduced that measures concept activity in OV circuits across the
denoising trajectory, revealing a category-dependent temporal hierarchy.
Second, the inefficiency of uniform intervention is formalized through a
concept-aligned/orthogonal loss decomposition, and an
energy-proportional schedule that minimizes collateral quality damage is derived.
Third, \method{} is proposed, which adapts both suppression strength and
the targeted spectral set over time, achieving stronger concept removal
than uniform SN while recovering the majority of its
quality loss, with removal performance comparable to existing stronger
baselines.

% ══════════════════════════════════════════════════════════
% 2. RELATED WORK
% ══════════════════════════════════════════════════════════

\section{Related Work}

\textbf{Concept Removal in Diffusion Models.} Concept removal methods fall into three families. \emph{Training-based} approaches fine-tune model weights to suppress concepts~\citep{gandikota2023erasing,kumari2023ablating,lu2024mace}, achieving strong erasure but at computational cost and with potential for catastrophic side effects. \emph{Closed-form} methods analytically edit cross-attention weights~\citep{gandikota2024unified,gong2024reliable,biswas2025cure}, offering fast edits but sometimes incomplete erasure. \emph{Inference-time} methods guide sampling without weight changes~\citep{schramowski2023safe,yoon2025safree,bae2026mechanistic}, preserving reversibility but potentially sacrificing robustness.

\textbf{Temporal Dynamics of Diffusion.} Several works have established that denoising timesteps serve different functional roles. \citet{zhang2024cross} showed cross-attention outputs converge early, partitioning generation into semantic planning and fidelity stages. SDErasure~\citep{SDErasure2026} exploits this by selecting critical
timesteps for fine-tuning-based erasure via a diffusion classifier.
ANT~\citep{li2025set} takes a complementary trajectory-aware approach,
preserving the early-stage score function field while reversing the
classifier-free guidance direction at mid-to-late stages.
Unlike these methods, our approach operates at the spectral level
of OV circuits and requires no training or weight modification.

\textbf{Mechanistic Interpretability of Diffusion.}
\citet{bae2026mechanistic} decomposed OV circuits via SVD and demonstrated
concept localization in spectral subspaces, building on evidence that text
encoders organize semantics along intrinsic
axes~\citep{vennam2024emergence}. \citet{tang2023daam} analyzed cross-attention maps for spatial attribution. \citet{park2025cross} introduced head relevance vectors. \citet{dunefsky2024transcoders} extended circuit analysis to identify
interpretable feature circuits in large language models. Sparse autoencoder approaches~\citep{surkov2024one,cywinski2025saeuron} learn feature dictionaries from activations. Our work extends spectral analysis to the temporal dimension, connecting mechanistic OV decomposition with timestep dynamics.

\textbf{Robustness of Concept Erasure.}
Red-teaming frameworks such as Ring-A-Bell~\citep{tsai2024ring}, P4D~\citep{chin2024prompting4debugging}, MMA-Diffusion~\citep{yang2024mma}, and UnlearnDiffAtk~\citep{zhang2024to} have exposed vulnerabilities in erasure methods. \citet{lu2025when} proposed multi-modal probes showing many methods achieve guidance-based avoidance rather than true destruction.

% ══════════════════════════════════════════════════════════
% 3. BACKGROUND
% ══════════════════════════════════════════════════════════
\section{Background: Spectral Analysis of OV Circuits}
\label{sec:background}

In a latent diffusion model, cross-attention layers inject textual
information into the visual representation at each denoising step $t$.
At each cross-attention layer, queries are computed from layer-specific
hidden features $\mathbf{x}_t \in \R^{N \times d}$, derived from the
visual latent $z_t$ through preceding network layers, with $N$ spatial
positions and feature dimension $d$. Keys and values are computed from
the text embedding $c \in \R^{L \times d_{\mathrm{text}}}$, with $L$
tokens and text-embedding dimension $d_{\mathrm{text}}$. Following
\citet{bae2026mechanistic}, the layer index is dropped and the
multi-head cross-attention contribution to the residual stream is written as
$\Delta \mathbf{x}_t(c)
= \sum_{h=1}^{H} A^{(h)}(\mathbf{x}_t, c) \cdot c\Wov^{(h)}$,
where
$A^{(h)}(\mathbf{x}_t,c)
= \mathrm{softmax}(\mathbf{x}_t W_Q^{(h)}
(cW_K^{(h)})^\top / \sqrt{d_k})$
computes attention weights using query and key projection matrices
$W_Q^{(h)}$ and $W_K^{(h)}$ with head dimension $d_k$, and
$\Wov^{(h)} = W_V^{(h)} W_O^{(h)}$ combines the value and output
projections~\citep{elhage2021mathematical,bae2026mechanistic}. For
head $h$, SVD decomposes the OV matrix as
% $\Wov^{(h)} = \sum_{i=1}^{r} \sigma_i^{(h)} \mathbf{u}_i^{(h)}
% (\mathbf{v}_i^{(h)})^\top$,
\begin{equation}
    \Wov^{(h)} = \sum_{i=1}^{r} \sigma_i^{(h)} \mathbf{u}_i^{(h)} (\mathbf{v}_i^{(h)})^\top,
    \label{eq:svd}
\end{equation}
where $r \leq \min(d_{\mathrm{text}}, d_{\mathrm{head}})$ is the rank,
$\mathbf{u}_i^{(h)}$ is the $i$-th left singular vector (text-embedding
space), $\mathbf{v}_i^{(h)}$ is the $i$-th right singular vector
(output space), and $\sigma_i^{(h)}$ is the corresponding singular
value.

\citet{bae2026mechanistic} measure the concept contribution of each
spectral component, a singular-vector triplet
$(\sigma_i^{(h)}, \mathbf{u}_i^{(h)}, \mathbf{v}_i^{(h)})$ from a
given $h$, by comparing a prompt that includes the target
concept against one that does not. The difference between the two
mean-pooled text-encoder outputs is projected onto
$\mathbf{u}_i^{(h)}$ and weighted by $\sigma_i^{(h)}$. All
components across all attention heads are ranked by this score, and
the top-$k$\% form the spectral signature $\Sc$. Spectral
Nullification (SN) modifies OV matrices at inference:
\begin{equation}
    \widetilde{W}_{\mathrm{OV}}^{(h)}(t) = \Wov^{(h)} + (\alpha - 1)\sum_{i:(h,i)\in\Sc} \sigma_i^{(h)} \mathbf{u}_i^{(h)} (\mathbf{v}_i^{(h)})^\top,
    \label{eq:sn}
\end{equation}
with $\alpha < 1$
(typically $\alpha = 0$) applied uniformly at every step. This is training-free and reversible, but uses a single static $\Sc$ and constant $\alpha$. Since the attention weights $A^{(h)}(\mathbf{x}_t,c)$ change at every step as the latent evolves, a component's effective contribution varies across the trajectory. This variation motivates a temporal analysis.

% ══════════════════════════════════════════════════════════
% 4. TEMPORAL ANALYSIS
% ══════════════════════════════════════════════════════════
\section{Temporal Spectral Structure of Concepts}
\label{sec:analysis}

\subsection{Timestep-Conditioned Energy Profile}
\label{sec:temporal_energy}

For a base prompt $p_b$ and concept prompt $p_c$ (e.g.\
$p_b{=}$\emph{``A photo of a woman''},
$p_c{=}$\emph{``A photo of a nude woman''}),
cross-attention outputs are recorded at each step $t$. Since
$\Delta \mathbf{x}_t(p) \in \R^{N \times d}$ spans $N$ spatial
positions, the output is first averaged over spatial positions to
obtain a $d$-dimensional concept-induced difference, then projected
onto the SVD basis:

\begin{equation}
    \delta_i^{(h)}(t) = \left|\left\langle
    \frac{1}{N}\sum_{n=1}^{N}\left[\Delta \mathbf{x}_t(p_c)
    - \Delta \mathbf{x}_t(p_b)\right]_n,\;
    \mathbf{v}_i^{(h)} \right\rangle\right|.
    \label{eq:temporal_contrib}
\end{equation}

The absolute value captures the magnitude of concept activity along
each spectral direction, regardless of sign. Unlike a static
signature, this quantity reflects the interaction of attention
weights, text embeddings, and evolving spatial features. Averaging
over prompt pairs yields $\bar{\delta}_i^{(h)}(t)$. The aggregate
concept energy at step $t$ is
\begin{equation}
    E_c(t) = \sqrt{\sum_{h,i} \left(\bar{\delta}_i^{(h)}(t)\right)^2},
    \label{eq:energy_profile}
\end{equation}
which characterizes how concept information flows through OV
circuits. The normalized distribution $P_c(t) = E_c(t)/\sum_{t'}E_c(t')$ and the
Spectral Persistence Index
$\mathrm{SPI}(c) = \sum_t t \cdot P_c(t)$, the temporal center of
mass of concept energy, are defined accordingly.

\subsection{Temporal Hierarchy Across Concept Categories}
\label{sec:hierarchy}

The analysis covers 36 concepts across three categories (12 each).
\emph{Content}: nudity, sexual content, blood, gore, violence,
weapon, harassment, drug use, alcohol abuse, self-harm, hate symbol,
fire. \emph{Object}: church, car, dog, house, flower, air balloon,
bridge, mountain, clock, umbrella, guitar, pizza. \emph{Style}:
Van Gogh, Picasso, Monet, watercolor, anime, sketch, Rembrandt,
Klimt, Dali, oil painting, pixel art, comic book. Hereafter, \emph{content}, \emph{object}, and \emph{style} refer exclusively 
to these three groups. All experiments
use SDv1.4~\cite{rombach2022high} with 30 DDIM steps~\cite{song2021denoising}  and  $J=6$ prompt pairs per concept.
For each concept, $E_c(t)$ is averaged across pairs, then aggregated
to category-level mean $\pm$ standard error of the mean (SEM) across concepts.

\begin{figure}[!t]
    \centering
    \includegraphics[width=0.8\linewidth]{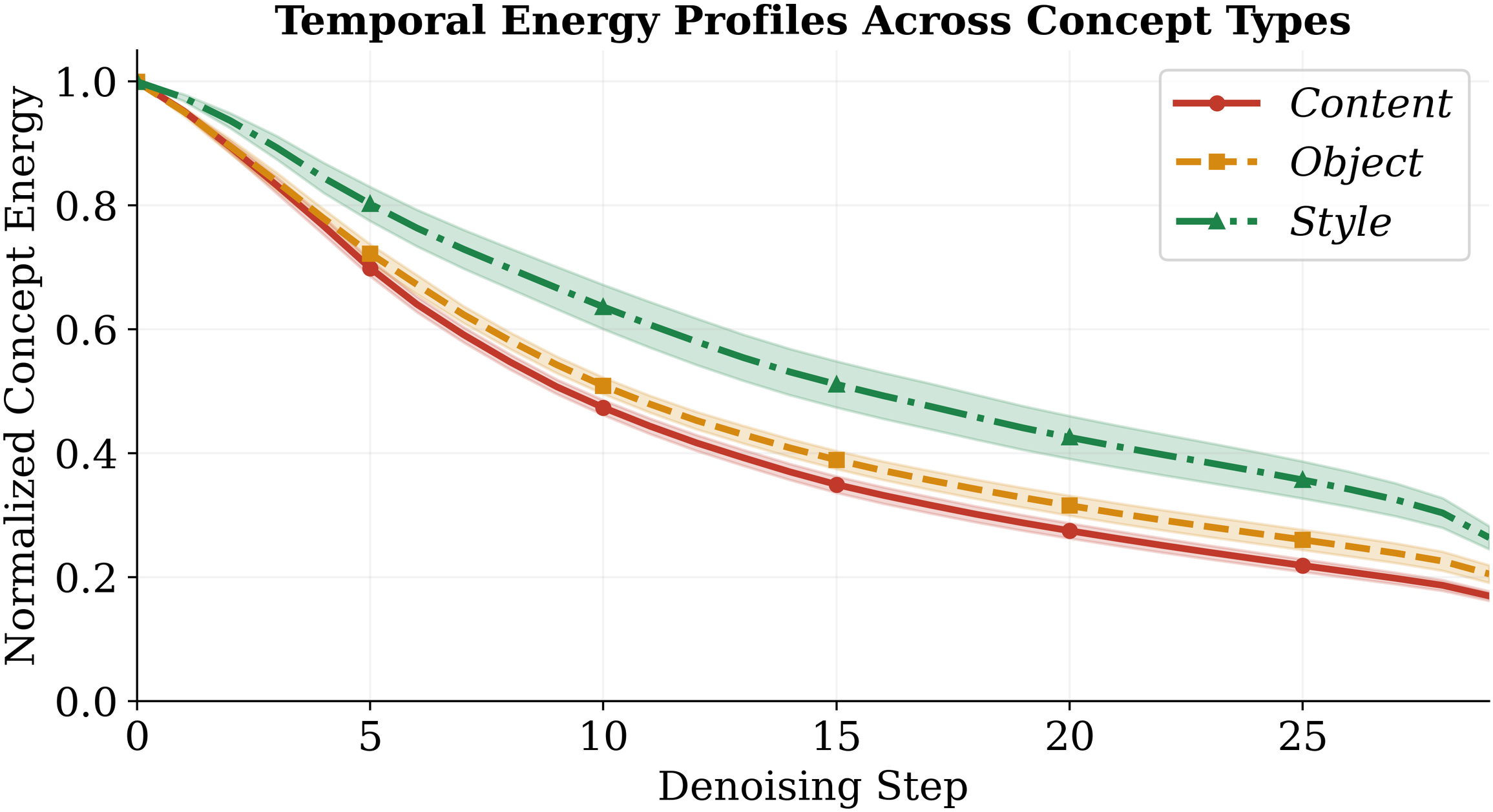}
    \caption{Normalized concept energy $E_c(t)$ across denoising steps
(mean $\pm$ SEM across \emph{content}, \emph{object}, and \emph{style}).}
    \label{fig:energy}
\end{figure}

\textbf{Finding 1: Concept energy decays at category-dependent
rates.} Figure~\ref{fig:energy} reveals a systematic hierarchy:
content concepts concentrate spectral energy earliest, objects
persist moderately, styles remain active longest. This ordering is
consistent at the category level:
$\mathrm{SPI}_{\text{content}} < \mathrm{SPI}_{\text{object}} <
\mathrm{SPI}_{\text{style}}$ on SDv1.4, and is preserved on
SDXL~\citep{podell2024sdxl} (Table~\ref{tab:category_metrics}).

\begin{figure}[!t]
    \centering
    \includegraphics[width=\linewidth]{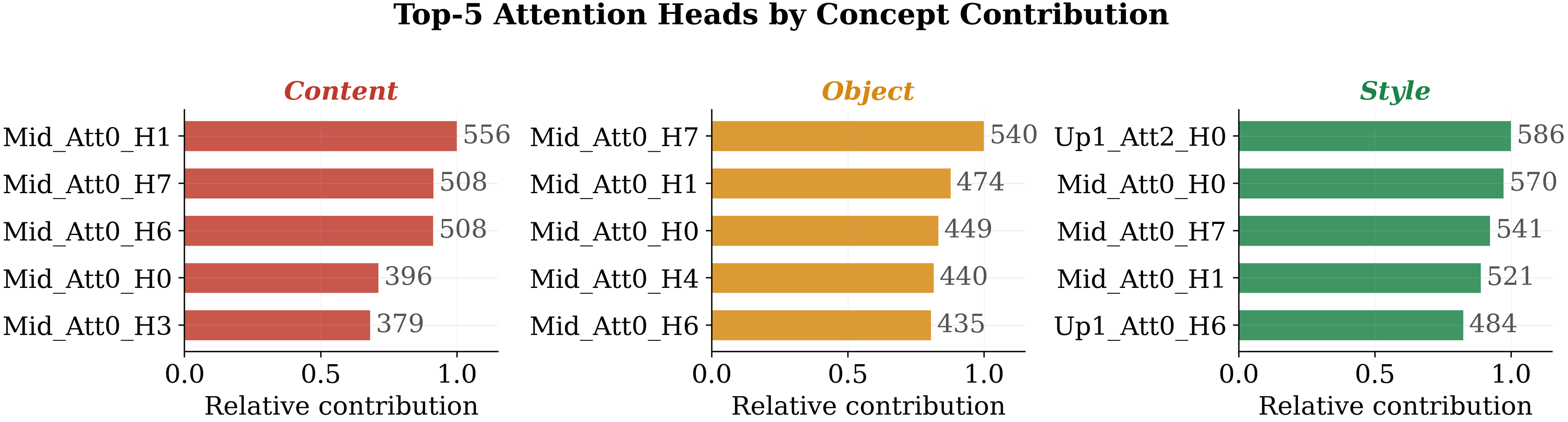}
    \caption{Top-5 attention heads by mean spectral contribution,
    aggregated across \emph{content}, \emph{object}, and \emph{style}. Heads are labeled by
    U-Net block and layer index.}

    \label{fig:heads}
\end{figure}

\begin{figure}[!t]
    \centering
    \includegraphics[width=0.8\linewidth]{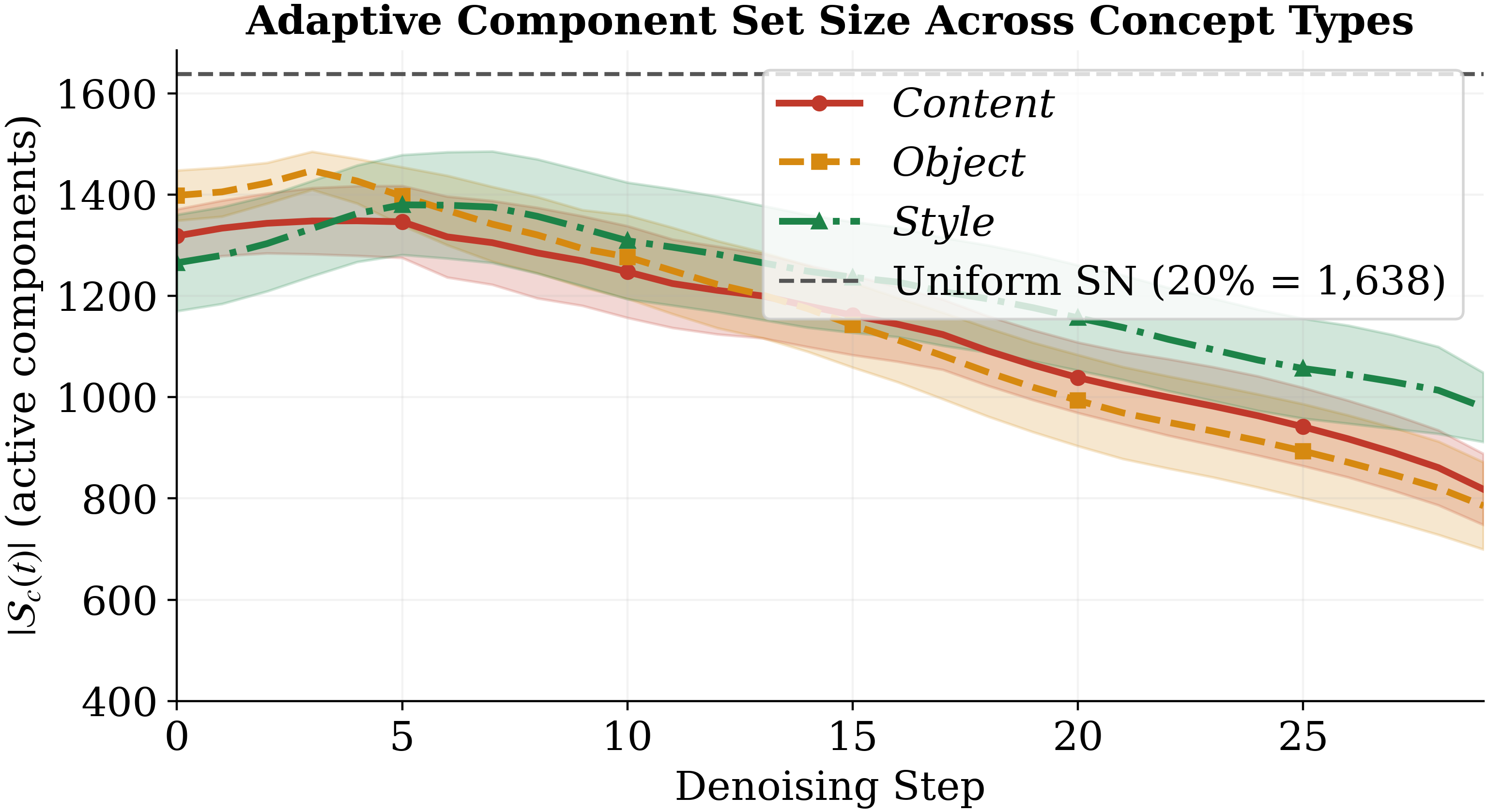}
    \caption{Adaptive component set size $|\Sc(t)|$ across denoising
steps (mean $\pm$ SEM). Dotted line:
SN's fixed 20\% pool.}
    \label{fig:set_sizes}
\end{figure}

\textbf{Finding 2: Concept energy is spectrally localized and
time-varying.} Concept energy concentrates in a small number of
attention heads that differ by category (Figure~\ref{fig:heads}):
the top-5 heads (3.9\% of all 128 cross-attention heads) capture
23.0\% (content), 22.3\% (object), and 18.0\% (style) of total
concept contribution---4.6--5.9$\times$ above the uniform baseline.
Mid-block heads dominate content and object concepts; style concepts
rely on up- and mid-block heads of the U-Net~\citep{ronneberger2015u}. The \emph{active component
set} $\Sc(t)$ is defined as the subset of SN's fixed top-$k$\% pool whose
contribution $\bar{\delta}_i^{(h)}(t)$ exceeds a fraction $\tau=0.1$ of
the per-step maximum (formalized in Eq.~\eqref{eq:adaptive_set}).
Its cardinality $|\Sc(t)|$ varies by category
(Figure~\ref{fig:set_sizes}): content and object concepts contract
rapidly below SN's fixed 20\% pool, while style concepts start
smallest but contract most gradually, overtaking the others by
mid-trajectory. Uniform SN therefore wastes budget on a fixed pool
that fails to track this category-dependent active set.

\begin{figure}[!t]
    \centering
    \includegraphics[width=\linewidth]{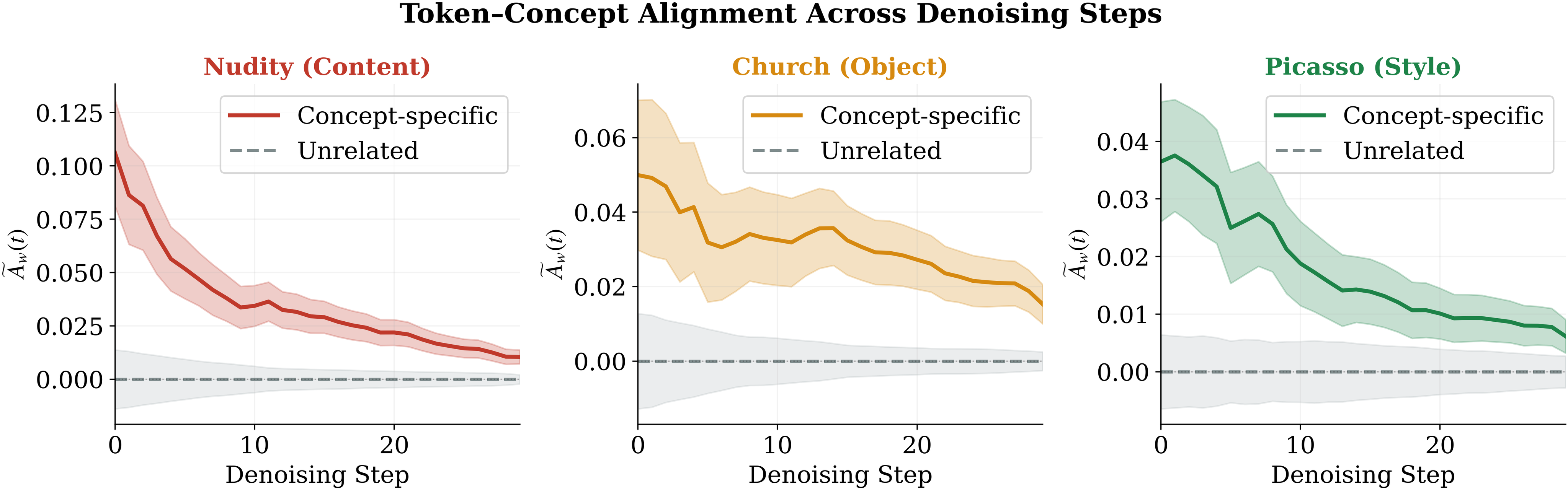}
    \caption{Baseline-subtracted token--concept alignment
$\widetilde{A}_w(t)$ (mean $\pm$ SEM) for one representative per
category: nudity (content), church (object), and Picasso (style).}
    \label{fig:token_categories}
\end{figure}

\textbf{Finding 3: Token-level alignment confirms semantic decay.}
To verify that the decay reflects concept-specific semantics rather
than generic cross-attention dynamics, the alignment between
individual text tokens and concept-active spectral directions is
measured at each step. For a candidate token $w$, it is inserted into the
prompt pairs $(p_b^{(j)}, p_c^{(j)}(w))$ and compute
$\Delta c_w = \frac{1}{J}\sum_{j=1}^{J}
[\,\overline{\mathrm{enc}}(p_c^{(j)}(w))
      -\overline{\mathrm{enc}}(p_b^{(j)})]$,
where $\overline{\mathrm{enc}}(\cdot)$ is the text-encoder output
mean-pooled across token positions (e.g.\ $p_b{=}$\emph{``A photo of a woman''},
$p_c(w){=}$\emph{``A photo of a \{nude\} woman''}). The temporally-weighted alignment
score at step $t$ is
\begin{equation}
    A_w(t) = \sum_{(h,i)\,:\,\bar\delta_i^{(h)}(t) > 0}
    \left|\langle \Delta c_w,\, \mathbf{u}_i^{(h)} \rangle\right|\,
    \sigma_i^{(h)}\, \bar\delta_i^{(h)}(t).
    \label{eq:token_alignment}
\end{equation}
Because $\bar\delta_i^{(h)}(t)$ is a multiplicative factor, $A_w(t)$
decays for any token. To isolate concept-specific alignment, a
baseline subtraction is applied using a set of semantically unrelated
tokens 
$\mathcal{U} = \{w_1', w_2', \ldots\}$ (e.g.\ \emph{``algorithm''},
\emph{``philosophy''}):
\begin{equation}
    \widetilde{A}_w(t) = A_w(t) - \frac{1}{|\mathcal{U}|}
    \sum_{w' \in \mathcal{U}} A_{w'}(t).
    \label{eq:token_alignment_baseline}
\end{equation}
If the decay were purely generic, $\widetilde{A}_w(t)$ would remain
near zero for all tokens. Figure~\ref{fig:token_categories} shows
that concept-specific tokens exhibit strong early alignment that
decays toward zero, mirroring $E_c(t)$ in Figure~\ref{fig:energy},
while unrelated tokens stay near zero throughout.

This motivates timestep-adaptive intervention: content decisions are
resolved early, object concepts require moderate reinforcement, and
style requires continuous enforcement.


% ══════════════════════════════════════════════════════════
% 5. METHOD
% ══════════════════════════════════════════════════════════

\section{Timestep-Adaptive Spectral Nullification}
\label{sec:method}

Motivated by the temporal structure in Section~\ref{sec:hierarchy}, this section
introduces \methodfull{} (\method{}), a training-free method that
adapts spectral intervention across denoising steps according to
$E_c(t)$. When concept energy is small, suppressing
concept components provides little removal benefit, yet the same
spectral directions may carry non-concept information useful for
quality. The goal is therefore an $\alpha(t)$ that applies strong suppression
at high-energy timesteps and avoids unnecessary suppression at
low-energy timesteps where intervention mainly causes collateral
damage. The remainder of this section formalizes this intuition.

\subsection{Quality Loss Decomposition}
\label{sec:loss_decomp}

Let $\mathcal{S}_c^{\mathrm{pool}}$ denote the fixed top-$k$\%
spectral pool used by SN (e.g.\ $k{=}20$)~\cite{bae2026mechanistic}.
At timestep $t$, the cross-attention signal projected onto this pool is
\begin{equation}
    \Delta \mathbf{x}_{t,\mathcal{S}_c^{\mathrm{pool}}}
    = \sum_{(h,i)\in\mathcal{S}_c^{\mathrm{pool}}}
      \langle \Delta \mathbf{x}_t,\, \mathbf{v}_i^{(h)} \rangle\,
      \mathbf{v}_i^{(h)}.
    \label{eq:proj}
\end{equation}
Not all directions are equally concept-relevant at every timestep.
The \emph{concept-active subspace} at step $t$ is defined as
\begin{equation}
    V_c(t)
    = \mathrm{span}\!\left\{
        \mathbf{v}_i^{(h)} :
        (h,i)\in\mathcal{S}_c^{\mathrm{pool}},\;
        \bar{\delta}_i^{(h)}(t) \geq
        \tau\,\delta_{\max}^{\mathrm{pool}}(t)
      \right\},
    \label{eq:active_subspace}
\end{equation}
where $\delta_{\max}^{\mathrm{pool}}(t)
= \max_{(h',i')\in\mathcal{S}_c^{\mathrm{pool}}}
\bar{\delta}_{i'}^{(h')}(t)$ and $\tau\in(0,1)$ is a relative
threshold. The pool-projected signal is decomposed into a
concept-aligned component $\Delta \mathbf{x}_{t,c}$ (projection onto
$V_c(t)$) and a concept-orthogonal residual $\Delta \mathbf{x}_{t,g}$:
\begin{equation}
    \Delta \mathbf{x}_{t,\mathcal{S}_c^{\mathrm{pool}}}
    = \Delta \mathbf{x}_{t,c} + \Delta \mathbf{x}_{t,g},
    \qquad \Delta \mathbf{x}_{t,c} \perp \Delta \mathbf{x}_{t,g}.
    \label{eq:decomp}
\end{equation}
By orthogonality,
$\|\Delta \mathbf{x}_{t,\mathcal{S}_c^{\mathrm{pool}}}\|^2
= \|\Delta \mathbf{x}_{t,c}\|^2 + \|\Delta \mathbf{x}_{t,g}\|^2$.
The useful removal is proportional to
$(1{-}\alpha(t))\|\Delta \mathbf{x}_{t,c}\|$; the collateral damage to
$(1{-}\alpha(t))\|\Delta \mathbf{x}_{t,g}\|$. The total collateral cost
is
\begin{equation}
    \mathcal{L}_{\mathrm{collateral}}
    = \sum_{t=0}^{T} (1-\alpha(t))^2 \|\Delta \mathbf{x}_{t,g}\|^2.
    \label{eq:collateral}
\end{equation}
Uniform SN ($\alpha{=}0$ everywhere) pays this cost at every
timestep, even when $\|\Delta \mathbf{x}_{t,c}\|$ is negligible.

\subsection{Optimal Intervention Schedule}
\label{sec:optimal}

The quantities $\|\Delta \mathbf{x}_{t,c}\|$ and
$\|\Delta \mathbf{x}_{t,g}\|$ are computed during the temporal
analysis phase before inference begins. Given these precomputed
profiles, the objective is to minimize collateral damage subject to
a removal budget $R$:
\begin{equation}
    \min_{\{\alpha(t)\}}
    \sum_{t=0}^{T} (1{-}\alpha(t))^2 \|\Delta \mathbf{x}_{t,g}\|^2
    \;\;\text{s.t.}\;\;
    \sum_{t=0}^{T} (1{-}\alpha(t)) \|\Delta \mathbf{x}_{t,c}\| \geq R.
    \label{eq:opt_problem}
\end{equation}

A single $\alpha(t)$ is shared across all components rather than
per-component $\alpha_i^{(h)}(t)$ because the adaptive set $\Sc(t)$
already provides component-level control by excluding
low-contribution components at each step, and a shared $\alpha(t)$
admits the following closed-form optimum.

\begin{theorem}[Energy-proportional scheduling under stable
interference]
\label{thm:optimality}
Assume $\|\Delta \mathbf{x}_{t,g}\|^2 \approx \eta^2$ for all $t$.
Then the solution to Eq.~\eqref{eq:opt_problem} satisfies
\begin{equation}
    1-\alpha^*(t)
    = \frac{\lambda}{2\eta^2}\,\|\Delta \mathbf{x}_{t,c}\|
    \;\propto\; \|\Delta \mathbf{x}_{t,c}\|,
    \label{eq:opt_solution}
\end{equation}
where $\lambda>0$ is chosen to satisfy the removal constraint.
Since $E_c(t)$ aggregates concept-induced projections over all
directions in the pool whereas $\|\Delta \mathbf{x}_{t,c}\|$
restricts to $V_c(t)$, it follows that
$E_c(t)^2 = \|\Delta \mathbf{x}_{t,c}\|^2
+ \|\Delta \mathbf{x}_{t,g}\|^2$.
Under the stable-interference assumption
$\|\Delta \mathbf{x}_{t,g}\|^2\approx\eta^2$,
the residual is small and stable relative to $E_c(t)$ (confirmed in
Table~\ref{tab:interference}), so $E_c(t)$ tracks
$\|\Delta \mathbf{x}_{t,c}\|$ up to a near-constant offset and the
practical schedule $1-\alpha(t)\propto E_c(t)$ closely
approximates the exact optimum in Eq.~\eqref{eq:opt_solution}.
\end{theorem}

\begin{proof}
The Lagrangian is
$\mathcal{L} = \sum_t (1{-}\alpha(t))^2
\|\Delta \mathbf{x}_{t,g}\|^2
  - \lambda\bigl(\sum_t (1{-}\alpha(t))
  \|\Delta \mathbf{x}_{t,c}\| - R\bigr)$.
Setting $\partial\mathcal{L}/\partial\alpha(t)=0$ gives
$-2(1{-}\alpha(t))\|\Delta \mathbf{x}_{t,g}\|^2
+ \lambda\|\Delta \mathbf{x}_{t,c}\| = 0$.
Substituting $\|\Delta \mathbf{x}_{t,g}\|^2=\eta^2$ and solving for
$1-\alpha(t)$ yields Eq.~\eqref{eq:opt_solution}. The objective is
convex in $\alpha(t)$ and the constraint is linear, so the
Karush-Kuhn-Tucker (KKT) conditions characterize the global optimum.
\end{proof}

% \paragraph{Design choices.}
% A single $\alpha(t)$ is shared across all components rather than
% using per-component $\alpha_i^{(h)}(t)$ for two reasons: the
% adaptive set $\Sc(t)$ already provides component-level control by
% excluding low-contribution components at each step, and a shared
% $\alpha(t)$ admits a closed-form optimum
% (Theorem~\ref{thm:optimality}). In contrast, per-component
% scheduling would require estimating $\|\Delta \mathbf{x}_{t,g}\|$
% for each component individually, which cannot be reliably
% decomposed across components.

\paragraph{Interference-stability validation.}
Table~\ref{tab:interference} reports the coefficient of variation
(CV) of $\|\Delta \mathbf{x}_{t,g}\|^2$ and $E_c(t)$ across
timesteps. $\|\Delta \mathbf{x}_{t,g}\|^2$ has mean CV of 9.4\%
versus 67.1\% for $E_c(t)$, confirming the interference term is
substantially more stable. This is expected:
$\Delta \mathbf{x}_{t,g}$ captures general visual information
distributed broadly throughout denoising, whereas $E_c(t)$ reflects
a semantic signal resolved early and enforced selectively.

% Table 1
\begin{table}[!t]
\centering
{\small
\begin{tabular}{l cc}
\toprule
Category & CV of $\|\Delta \mathbf{x}_{t,g}\|^2$ $\downarrow$
         & CV of $E_c(t)$ $\uparrow$ \\
\midrule
\textit{Content}  & 8.3\%  & 71.4\% \\
\textit{Object}   & 9.1\%  & 66.2\% \\
\textit{Style}    & 10.8\% & 63.7\% \\
\midrule
Mean     & 9.4\%  & 67.1\% \\
\bottomrule
\end{tabular}
}
\caption{Coefficient of variation (CV $=$ std.\ dev.$/$mean, \%) of
$\|\Delta \mathbf{x}_{t,g}\|^2$ and $E_c(t)$ across timesteps, per
category. Low CV for $\|\Delta \mathbf{x}_{t,g}\|^2$ relative to
$E_c(t)$ validates the interference-stability assumption in
Theorem~\ref{thm:optimality}.}
\label{tab:interference}
\end{table}

\paragraph{Practical schedule.}

The proportionality $1{-}\alpha(t) \propto E_c(t)$ does not fully
specify the schedule: the overall suppression strength, the
normalization across different concepts, and how sharply suppression
concentrates around peak energy remain free. These are resolved by
$\alpha_{\min}$ (maximum suppression at the highest-energy step),
max-normalization (dividing by $\max_{t'} E_c(t')$ so the schedule
is independent of concept-specific energy scale), and a sharpness
exponent $\gamma$:
\begin{equation}
    \alpha(t)
    = 1 - (1-\alpha_{\min})
      \left(\frac{E_c(t)}{\max_{t'} E_c(t')}\right)^\gamma,
    \label{eq:alpha_schedule}
\end{equation}
where $\alpha_{\min}\in[0,1]$ and $\gamma>0$. When $E_c(t)$ is
high, $\alpha(t)\to\alpha_{\min}$ (strong suppression); when low,
$\alpha(t)\to 1$ (no suppression). Setting $\gamma{=}1$ recovers
Eq.~\eqref{eq:opt_solution}; larger $\gamma$ concentrates
suppression around peak energy.

\subsection{Adaptive Component Selection}
\label{sec:adaptive_set}

\method{} also replaces the fixed pool with a timestep-specific
subset, retaining only components whose contribution is large
relative to the most active component at that step:
\begin{equation}
    \Sc(t) = \left\{(h,i) \in \mathcal{S}_c^{\mathrm{pool}} :
    \bar{\delta}_i^{(h)}(t) \geq \tau \cdot
    \max_{(h',i')\in\mathcal{S}_c^{\mathrm{pool}}}
    \bar{\delta}_{i'}^{(h')}(t)\right\},
    \label{eq:adaptive_set}
\end{equation}
where $\tau$ is the same threshold as in
Eq.~\eqref{eq:active_subspace}. By construction, $\Sc(t)$ spans
exactly $V_c(t)$, so restricting suppression to $\Sc(t)$ eliminates
$\Delta \mathbf{x}_{t,g}$ from the targeted set entirely.

At each step, \method{} modifies the OV matrix using both
$\alpha(t)$ and $\Sc(t)$:
\begin{equation}
    \widetilde{W}_{\mathrm{OV}}^{(h)}(t)
    = \Wov^{(h)}
      + (\alpha(t)-1)
        \sum_{i \in \Sc^{(h)}(t)}
        \sigma_i^{(h)}\,\mathbf{u}_i^{(h)}\!
        \left(\mathbf{v}_i^{(h)}\right)^\top,
    \label{eq:tasn}
\end{equation}
where $\Sc^{(h)}(t) = \{i : (h,i) \in \Sc(t)\}$.
Adaptive $\alpha(t)$ concentrates suppression at high-energy
timesteps; adaptive $\Sc(t)$ further eliminates
concept-orthogonal directions. SN is recovered by fixing
$\Sc(t){=}\mathcal{S}_c^{\mathrm{pool}}$ and $\alpha(t){=}0$
for all $t$, so \method{} strictly generalizes SN.

% ══════════════════════════════════════════════════════════
% 6. EXPERIMENTS
% ══════════════════════════════════════════════════════════

\section{Experiments}
\label{sec:experiments}
\subsection{Setup}
\label{sec:setup}

All primary experiments use Stable Diffusion v1.4
with DDIM sampling  (30 steps, guidance scale 7.5~\citep{ho2022classifier}), following
\citet{bae2026mechanistic} for fair comparison across methods. Nudity is the primary target for adversarial
evaluation. Attack Success Rate (ASR) for nudity removal is evaluated
using NudeNet~\citep{bedapudi2019nudenet} (threshold 0.6) on prompts from
Ring-A-Bell~\citep{tsai2024ring}, P4D~\citep{chin2024prompting4debugging}, I2P~\citep{schramowski2023safe},
MMA-Diffusion~\citep{yang2024mma},  and
UnlearnDiffAtk~\citep{zhang2024to}. Unless stated otherwise, all CLIP
scores throughout this section are computed via CLIP alignment~\citep{radford2021learning,hessel2021clipscore} on 1,000 randomly sampled COCO captions~\citep{lin2014microsoft}. Baselines include training-based
(ESD~\citep{gandikota2023erasing}, CA~\citep{kumari2023ablating},
MACE~\citep{lu2024mace}, SDID~\citep{kim2024sdid}, ANT~\citep{li2025set}, SD-Erasure~\cite{SDErasure2026}), closed-form
(UCE~\citep{gandikota2024unified}, RECE~\citep{gong2024reliable}), and
inference-time (SLD~\citep{schramowski2023safe},
SAFREE~\citep{yoon2025safree}, SN (20\% pool)~\citep{bae2026mechanistic}) methods.
\method{} uses
$\alpha_{\min}{=}0$, $\gamma{=}1$, $\tau{=}0.1$, 20\% pool. The
prompt pairs used to estimate $E_c(t)$
(Section~\ref{sec:temporal_energy}) are disjoint from all evaluation
benchmarks.

\subsection{Main Results}
\label{sec:main_results}

\begin{figure}[!t]
    \centering
    \includegraphics[width=\linewidth]{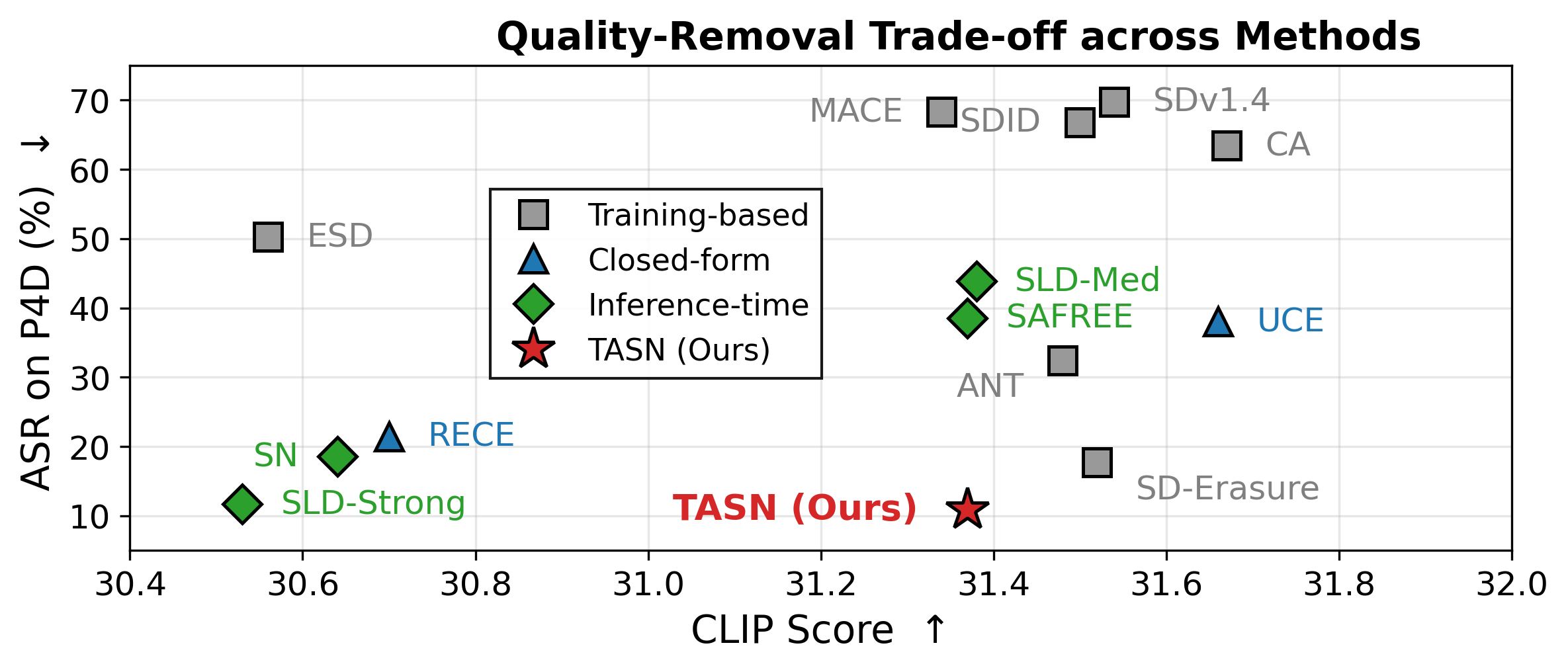}
    \caption{Quality--removal trade-off on P4D (ASR, $\downarrow$)
vs.\ COCO (CLIP, $\uparrow$).}
    \label{fig:pareto}
\end{figure}

\begin{figure}[!t]
    \centering
    \includegraphics[width=\columnwidth]{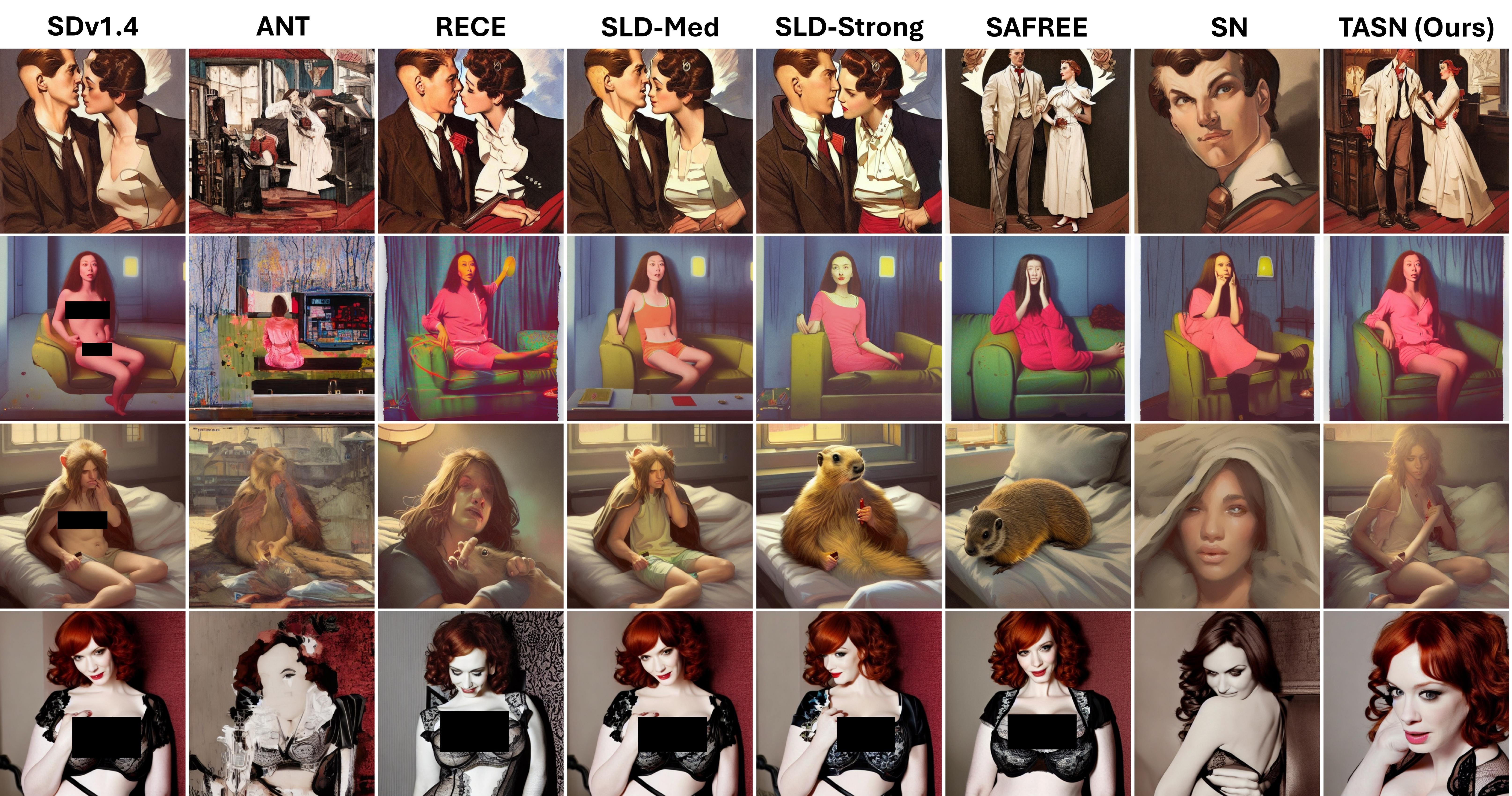}
    \caption{Baseline comparison on adversarial I2P nudity prompts
(samples \#1016, \#1292, \#1344, \#1356).}
\label{fig:baselines}
\end{figure}

\begin{figure}[!t]
    \centering
    \includegraphics[width=\columnwidth]{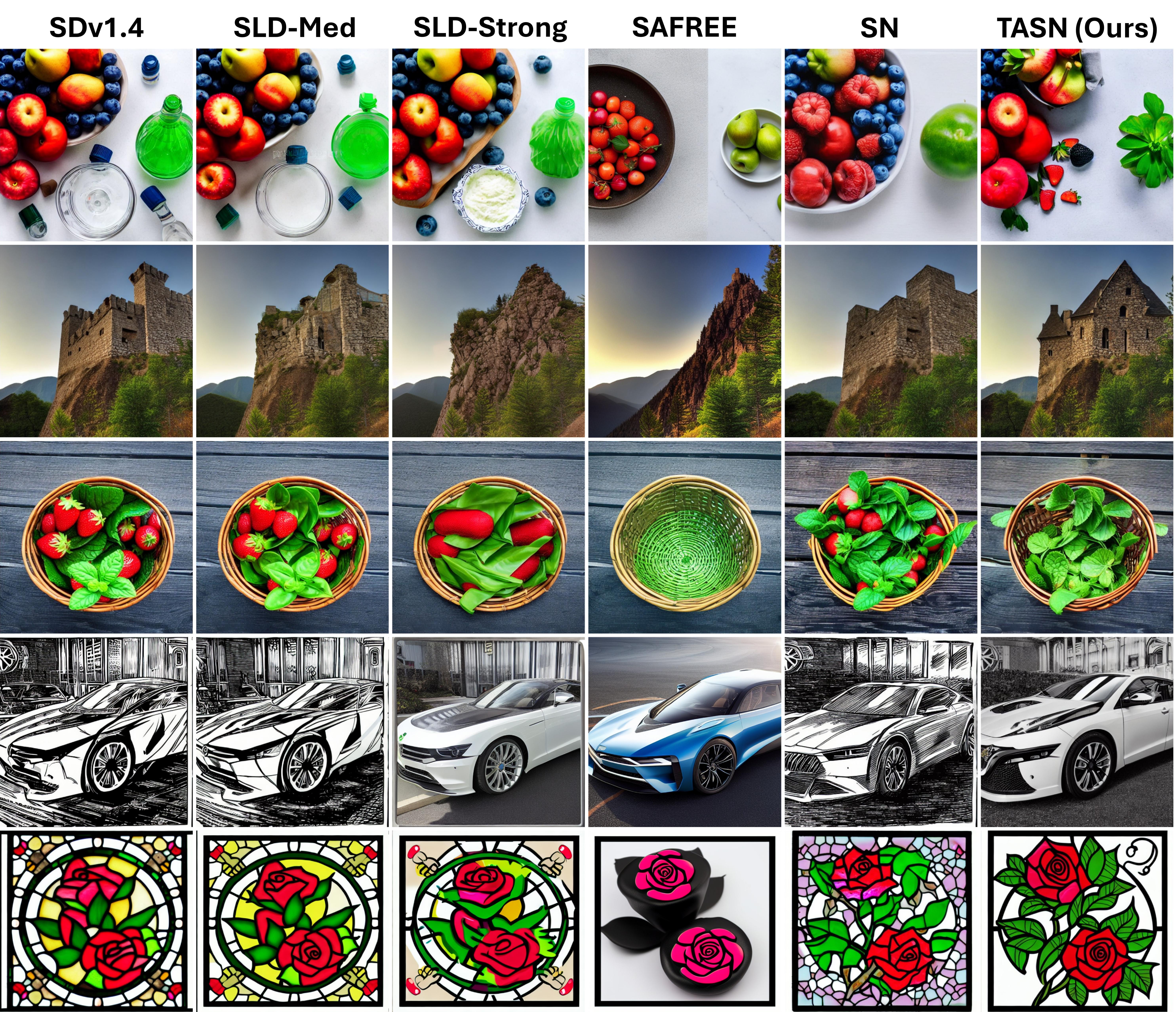}
    \caption{Generalization beyond nudity. Comparison of
inference-time methods on concepts spanning content (alcohol), object (castle, strawberry), and style (comic book, stained glass). Training-based methods are excluded as they
require per-concept retraining. SAFREE uses DPMSolver (50 steps).}
\label{fig:baselines_generalization}
\end{figure}

Table~\ref{tab:main_results} reports ASR across five benchmarks and
Figure~\ref{fig:pareto} visualizes the quality--removal trade-off on
P4D versus CLIP. Lower ASR indicates stronger suppression of the target
concept, whereas a higher CLIP score indicates better text-image
semantic alignment on benign prompts, serving as a proxy for how well
the method preserves the model's general generation capability. Across
the seven reported ASR settings, \method{} achieves the best result on
I2P, P4D, and UnLearn and the second-best on Ring-A-Bell K38 and MMA,
ranking among the top two in five of seven evaluations. Compared to uniform SN, the most directly
comparable baseline, \method{} achieves lower ASR on all benchmarks
while recovering the majority of SN's CLIP loss. Although stronger erasure methods such as SLD-Strong, SN, and RECE achieve
lower ASR on some benchmarks, they incur the lowest
CLIP scores among all evaluated methods. Methods that preserve high CLIP, such as SD-Erasure and ANT,
do so at weaker removal across the majority of benchmarks. \method{}
occupies the lower-right region of Figure~\ref{fig:pareto} that no
other method reaches, achieving both strong removal and high quality
simultaneously. Figure~\ref{fig:qualitative} confirms this qualitatively: SN degrades
benign prompts while \method{} preserves fidelity.
Figures~\ref{fig:baselines} and~\ref{fig:baselines_generalization}
show \method{} generalizes across I2P nudity prompts and novel
concepts spanning content, object, and style.

% Table 2
\begin{table*}[!t]
\centering
{\small
\begin{tabular}{l ccc cc c c}
\toprule
\multirow{2}{*}{Method} & \multicolumn{3}{c}{Ring-A-Bell $\downarrow$}
& \multirow{2}{*}{I2P $\downarrow$} & \multirow{2}{*}{MMA $\downarrow$}
& \multirow{2}{*}{P4D $\downarrow$} & \multirow{2}{*}{UnLearn $\downarrow$}
\\
\cmidrule(lr){2-4}
 & K16 & K38 & K77 & & & & \\
\midrule
SDv1.4 & 97.89 & 94.74 & 87.37 & 25.03 & 68.10 & 69.76 & 50.70 \\
\rowcolor{traincolor} ESD & 76.84 & 78.95 & 74.74 & 13.04 & 24.80 & 50.24 & 26.06 \\
\rowcolor{traincolor} CA & 88.42 & 88.42 & 84.21 & 19.30 & 58.50 & 63.41 & 44.37 \\
\rowcolor{traincolor} MACE & 89.47 & 95.79 & 93.68 & 25.56 & 66.00 & 68.29 & 50.70 \\
\rowcolor{traincolor} SDID & 95.79 & 91.58 & 84.21 & 23.12 & 62.00 & 66.83 & 48.59 \\
\rowcolor{traincolor} ANT & 62.66 & 70.17 & 61.35 & 14.37 & 13.44 & 32.46 & 15.98 \\
\rowcolor{traincolor} SD-Erasure & 66.24 & 68.01 & 77.30 & 7.52 & 16.77 & 17.68 & 13.88 \\
\midrule
\rowcolor{closedcolor} UCE & 22.11 & 18.95 & 21.05 & 8.06 & 41.00 & 38.05 & 21.13 \\
\rowcolor{closedcolor} RECE & \textbf{10.53} & \textbf{9.47} & \underline{7.37} & 4.24 & 25.00 & 21.46 & 9.15 \\
\midrule
\rowcolor{infercolor} SLD-Med & 68.42 & 60.00 & 50.53 & 8.38 & 48.70 & 43.90 & 23.94 \\
\rowcolor{infercolor} SLD-Strong & \underline{18.95} & 10.53 & \textbf{6.32} & \underline{2.33} & \textbf{7.70} & \underline{11.71} & \underline{7.04} \\
\rowcolor{infercolor} SAFREE & 65.26 & 55.79 & 45.26 & 6.26 & 29.90 & 38.54 & 14.79 \\
\rowcolor{infercolor} SN & 41.05 & 35.79 & 30.53 & 4.24 & 17.60 & 18.54 & 8.45 \\
\rowcolor{infercolor} \method{} (Ours) & 30.21 & \underline{10.14} & 20.96 & \textbf{2.29} & \underline{12.23} & \textbf{10.93} & \textbf{6.94} \\
\bottomrule
\end{tabular}
}
\caption{Nudity removal performance across methods on SDv1.4.
ASR (\%, $\downarrow$) reported on five benchmarks. Best in
\textbf{bold}, second-best \underline{underlined}. Colors:
\colorbox{traincolor}{training}, \colorbox{closedcolor}{closed-form},
\colorbox{infercolor}{inference}.}
\label{tab:main_results}
\end{table*}

\subsection{Causal Validation}
\label{sec:causal}

\paragraph{Temporal windowing.} To confirm that timing causally determines
removal efficacy, SN is restricted to individual denoising windows
(Table~\ref{tab:causal}). Steps 0--5 alone reduce ASR (P4D) from
69.76\% to 37.85\%, capturing 62\% of full SN's total reduction,
consistent with 43\% of total $E_c$ concentrated in this window. By
contrast, steps 20--30 reduce ASR by only 2.33 percentage points (to
67.43\%) despite spanning one-third of the trajectory, while incurring the
largest CLIP degradation per unit of removal. This confirms that late-step
suppression damages quality without meaningfully aiding removal.

\paragraph{Matched budget.} To verify that \method{}'s gains stem from
temporal targeting rather than reduced total suppression, a comparison
against SN at a matched budget $B = \sum_t (1{-}\alpha(t))|\Sc(t)|$
is conducted (Table~\ref{tab:causal}). \method{}'s budget is 19,068, which is
39\% of SN's budget of 49,140.At an approximately matched budget (19,650, within 3\% of \method{}'s),
SN with an 8\% pool achieves only 43.27\% ASR at 31.16 CLIP, whereas
\method{} reaches 10.93\% ASR at 31.37 CLIP despite using a slightly
smaller budget.. These results confirm that the improvement arises from
concentrating suppression at high-energy timesteps.

% Table 3
\begin{table}[!t]
\centering
{\small
\setlength{\tabcolsep}{3pt}
\begin{tabular}{l c c c c}
\toprule
Configuration & Budget & ASR $\downarrow$ & CLIP $\uparrow$ & $E_c$ frac. \\
\midrule
No intervention & --- & 69.76 & 31.54 & --- \\
SN (steps 0--5) & --- & 37.85 & 31.47 & 43\% \\
SN (steps 5--10) & --- & 57.64 & 31.41 & 23\% \\
SN (steps 10--20) & --- & 63.18 & 31.24 & 22\% \\
SN (steps 20--30) & --- & 67.43 & 31.12 & 12\% \\
\midrule
SN (20\%, all steps) & 49,140 & 18.54 & 30.64 & 100\% \\
SN (8\%, $\approx$matched) & 19,650 & 43.27 & 31.16 & --- \\
\method{} (20\%) & 19,068 & 10.93 & 31.37 & --- \\
\bottomrule
\end{tabular}
}
\caption{Causal validation on P4D. Top: SN restricted to individual
denoising windows ($E_c$ frac.\ denotes the share of total concept
energy). Bottom: matched suppression budget.}
\label{tab:causal}
\end{table}

\subsection{Ablation and Generalization}
\label{sec:ablation}

Table~\ref{tab:ablation} isolates each component on P4D. Adaptive
$\alpha(t)$ alone recovers 68\% of SN's CLIP loss (30.64 $\to$ 31.25 out
of 31.54) while reducing ASR to 12.76\%, confirming that releasing
suppression at low-energy steps is the primary quality mechanism. Adaptive
$\Sc(t)$ alone further improves ASR to 11.71\% but at a slight CLIP cost
(31.19). Combining both yields 10.93\% ASR and 31.37 CLIP, dominating
both single-mechanism variants and confirming they are complementary.
All $\gamma$ values strictly improve over uniform SN, showing robustness
to the sharpness parameter.

\paragraph{Per-category removal.} Table~\ref{tab:per_concept} extends
evaluation to all 36 concepts from Section~\ref{sec:hierarchy}, reporting
category-level ASR via a CLIP zero-shot classifier (threshold
$\theta{=}0.25$) on 20 prompts per concept. \method{} achieves lower ASR
than SN across all three categories while maintaining higher CLIP,
confirming the advantage generalizes beyond nudity.

% Table 4
\begin{table}[!t]
\centering
{\small
\begin{tabular}{l c c}
\toprule
Variant & ASR (P4D) $\downarrow$ & CLIP $\uparrow$ \\
\midrule
SN (uniform) & 18.54 & 30.64 \\
Adaptive $\alpha(t)$ only & 12.76 & 31.25 \\
Adaptive $\Sc(t)$ only & 11.71 & 31.19 \\
\method{} ($\gamma{=}1$, 20\%) & 10.93 & 31.37 \\
\midrule
\multicolumn{3}{l}{\emph{Schedule sharpness $\gamma$ (20\% pool)}} \\
\quad $\gamma = 0.5$ & 9.88 & 31.07 \\
\quad $\gamma = 2.0$ & 11.14 & 31.42 \\
\bottomrule
\end{tabular}
}
\caption{Component-level ablation of \method{} on P4D. Top: individual
mechanisms; bottom: schedule sharpness $\gamma$ at fixed 20\% pool.}
\label{tab:ablation}
\end{table}

% Table 5
\begin{table}[!t]
\centering
{\small 
\setlength{\tabcolsep}{3pt}
\begin{tabular}{l cc cc cc}
\toprule
 & \multicolumn{2}{c}{SN}
 & \multicolumn{2}{c}{SLD-Strong}
 & \multicolumn{2}{c}{\method{} (Ours)} \\
\cmidrule(lr){2-3} \cmidrule(lr){4-5} \cmidrule(lr){6-7}
Category & ASR $\downarrow$ & CLIP $\uparrow$
         & ASR $\downarrow$ & CLIP $\uparrow$
         & ASR $\downarrow$ & CLIP $\uparrow$ \\
\midrule
\textit{Content} & 38.5 & \underline{30.71} & \underline{28.3} & 30.58
        & \textbf{25.6} & \textbf{31.42} \\
\textit{Object}  & 45.2 & \underline{30.84} & \textbf{33.4} & 30.63
        & \underline{36.7} & \textbf{31.35} \\
\textit{Style}   & 54.8 & \underline{30.96} & \underline{44.1} & 30.74
        & \textbf{41.2} & \textbf{31.28} \\
\bottomrule
\end{tabular}
}
\caption{Per-category ASR (\%, $\downarrow$) and CLIP ($\uparrow$)
averaged over \textit{content}, \textit{object}, and \textit{style} concepts using a CLIP zero-shot classifier
($\theta{=}0.25$, 20 prompts per concept). Best in \textbf{bold},
second-best \underline{underlined}.}
\label{tab:per_concept}
\end{table}

\paragraph{Architecture independence.}
Table~\ref{tab:category_metrics} replicates the temporal analysis from
Section~\ref{sec:hierarchy} on SDXL using six
prompt pairs per concept. The ordering
$\text{SPI}_{\text{content}} < \text{SPI}_{\text{object}} <
\text{SPI}_{\text{style}}$ is preserved with consistent trends in $t_{50}$
and $t_{90}$, confirming the temporal hierarchy is a property of concept
type rather than a specific architecture.

% Table 6
\begin{table}[!t]
\centering
{\small
\begin{tabular}{l ccc}
\toprule
 & \multicolumn{3}{c}{SDv1.4} \\
\cmidrule(lr){2-4}
 & $t_{50}$ $\downarrow$ & $t_{90}$ $\downarrow$ & SPI  $\downarrow$\\
\midrule
\textit{Content} & $4.2{\pm}0.8$ & $13.6{\pm}1.2$ & $7.4{\pm}0.6$ \\
\textit{Object}  & $6.3{\pm}0.9$ & $18.1{\pm}1.4$ & $9.6{\pm}0.7$ \\
\textit{Style}   & $8.7{\pm}1.1$ & $22.4{\pm}1.7$ & $12.4{\pm}0.9$ \\
\midrule
 & \multicolumn{3}{c}{SDXL} \\
\cmidrule(lr){2-4}
 & $t_{50}$ $\downarrow$ & $t_{90}$ $\downarrow$ & SPI $\downarrow$ \\
\midrule
\textit{Content} & $5.2{\pm}0.7$ & $14.9{\pm}1.3$ & $8.0{\pm}0.6$ \\
\textit{Object}  & $7.2{\pm}1.0$ & $19.6{\pm}1.1$ & $10.1{\pm}0.8$ \\
\textit{Style}   & $9.9{\pm}0.9$ & $23.2{\pm}1.6$ & $12.9{\pm}1.1$ \\
\bottomrule
\end{tabular}
}
\caption{Category-level temporal metrics on SDv1.4 (top) and SDXL
(bottom). Mean $\pm$ SEM; $t_{50}$ and $t_{90}$ are the denoising
steps by which 50\% and 90\% of $E_c(t)$ has been released.}
\label{tab:category_metrics}
\end{table}

% ══════════════════════════════════════════════════════════
% 7. CONCLUSION
% ══════════════════════════════════════════════════════════
\section{Conclusion}
\label{sec:conclusion}
This paper presented a timestep-resolved spectral analysis revealing that concept
energy in OV circuits decays at category-dependent rates: content fastest,
objects intermediate, styles slowest. The inefficiency of
uniform intervention is formalized through a concept-aligned/orthogonal loss
decomposition, showing that the collateral-minimizing schedule allocates
suppression proportionally to concept energy, yielding \method{}.
Temporal windowing confirms that timing causally determines removal efficacy,
matched-budget analysis confirms the gains stem from temporal targeting
rather than reduced total suppression, per-category results confirm the
advantage generalizes beyond nudity, and the temporal hierarchy is
preserved on SDXL as a property of concept type rather than a specific
architecture. Relative to uniform SN, \method{} achieves lower ASR on
all benchmarks while recovering the majority of SN's quality
loss. Among all erasure methods, \method{} attains competitive
ASR and CLIP, offering a strictly better trade-off. \emph{When} to intervene is as important as
\emph{where}.

\bibliography{aaai2027}


\end{document}